\documentclass{llncs}
\pdfoutput=1
\usepackage[T1]{fontenc}
\usepackage{makeidx}

\usepackage{makeidx}
\usepackage{graphicx}
\usepackage{amssymb}
\usepackage{amsmath}
\usepackage{amscd}

\newcommand{\commentout}[1]{}

\newcommand{\introskip}{\vspace{3ex}}

\newcommand{\N}{\mathbb{N}}                    
\newcommand{\R}{\mathbb{R}}                    
\newcommand{\E}{\mathbb{E}}                    

\newcommand{\abs}[1]{\mathop{\left\lvert #1 \right\rvert}} 
\newcommand{\args}[1]{\mathop{\left( #1 \right)}} 
\newcommand{\inner}[1]{\mathop{\left\langle #1 \right\rangle}}
\newcommand{\norm}[1]{\mathop{\left\lVert #1 \right\rVert}}
\newcommand{\cbrace}[1]{\mathop{\left\{ #1 \right\}}}
\newcommand{\bracket}[1]{\mathop{\left[ #1 \right]}}

\newcommand{\argsS}[2]{\mathop{\left( #1 \right)#2}} 

\newcommand{\normS}[2]{\mathop{\left\lVert #1 \right\rVert#2}}

\DeclareMathOperator{\conv}{con}               

\DeclareMathOperator{\id}{id}                  
\newcommand{\T}{\mathop{\mathsf{T}}}           

\renewcommand{\S}[1]{{\mathcal{#1}}}           
\def\vec#1{\mathchoice{\mbox{\boldmath$\displaystyle#1$}}
{\mbox{\boldmath$\textstyle#1$}}
{\mbox{\boldmath$\scriptstyle#1$}}
{\mbox{\boldmath$\scriptscriptstyle#1$}}}

\renewenvironment{cases}{%
\left\{\begin{array}{c@{\quad : \quad}l}}%
{%
\end{array}\right.}

\newcounter{algorithm_counter}
{
\\[-1.5ex]
\rule{\textwidth}{\arrayrulewidth}
\end{footnotesize}
\end{minipage}
\setlength{\parindent}{\parindent}
}

{
\\[-1.5ex]
\rule{\textwidth}{\arrayrulewidth}
\end{footnotesize}
\end{minipage}
\setlength{\parindent}{\parindent}
}

\begin{document}
\title{Graph Quantization}
\titlerunning{}

\author{Brijnesh J.~Jain$^1$  \and Klaus Obermayer$^1$} 
\authorrunning{Brijnesh J.~Jain and Klaus Obermayer}  
\institute{$^1$Berlin Institute of Technology, Germany\\
\email{\{jbj|oby\}@cs.tu-berlin.de}}

\maketitle 

\begin{abstract}  
Vector quantization(VQ) is a lossy data compression technique from signal processing, which is restricted to feature vectors and therefore inapplicable for combinatorial structures. 
This contribution presents a theoretical foundation of graph quantization (GQ) that extends VQ to the domain of attributed graphs. We present the necessary Lloyd-Max conditions for optimality of a graph quantizer and consistency results for optimal GQ design based on empirical distortion measures and stochastic optimization. These results statistically justify existing clustering algorithms in the domain of graphs. The proposed approach provides a template of how to link structural pattern recognition methods other than GQ  to statistical pattern recognition. 
\end{abstract}

\section{Introduction}
Vector quantization is a classical technique from signal processing suitable for lossy data compression, density estimation, and prototype-based clustering  \cite{Duda00,Gersho92,Theodoridis09}. The problem of optimal vector quantizer design is to find a codebook consisting of a finite set of prototypes such that an expected distortion with respect to some (differentiable) distortion measure is minimized. 

Since the probability distribution of the input patterns is usually unknown, vector quantizer design techniques use empirical data. Extensively studied design techniques are, for example, k-means and simple competitive learning. The k-means algorithm is also commonly referred to as the Linde-Buzo-Gray (LBG) algorithm \cite{Linde80} the generalized Lloyd algorithm \cite{Lloyd57}. This algorithm is a local optimizer of the empirical sum-of-squared-error distortion without any global optimal or consistency guarantees. In contrast to k-means, competitive learning directly minimizes the expected distortion and is a consistent learner under very general conditions in the sense that it almost surely converges to a local optimal solution of the expected distortion.

One limitation of VQ is its restriction to patterns that are represented by vectors. For patterns that are more naturally represented by finite combinatorial structures, the theoretical framework of VQ as well its design techniques are no longer applicable. Examples of such structures include, for example, point patterns, strings, trees, and graphs arising from diverse application areas like proteomics, chemoinformatics, and computer vision.  

To overcome this limitation,  we generalize vector quantization to quantization of graphs. A number of graph quantizer design techniques for the purpose of prototype-based clustering have already been proposed. Examples include competitive learning algorithms in the domain of graphs \cite{Gold96b,Guenter02,Hagenbuchner03,Jain04,Jain08,Jain09b} and k-means as well as k-medoids algorithms  \cite{Ferrer07,Ferrer09,Jain04,Jain08,Jain09c,Schenker03,Schenker05}.  Related clustering method are presented in \cite{Bunke03,Lozano03,Torsello06}. Due to a lack of an appropriate theoretical framework, all these graph quantizer design techniques (or clustering methods) have been developed in order to minimize an empirical distortion function without justifying whether the solutions found are statistically consistent estimators of the true but unknown solutions. In addition, it is unclear whether the nearest neighbor and centroid condition, which are also referred to as the Lloyd-Max conditions, are necessary conditions for optimality. 

In this contribution, we propose graph quantization in a mathematically principled way as an extension of vector quantization, where we consider the graph edit distance as an underlying graph distortion measure. The key results of this contribution are consistency statements for estimators based on empirical distortion measures and estimators based on stochastic optimization. Furthermore, we prove that the Llyod-Max conditions are also necessary condition for optimal graph quantizers. In order to achieve the consistency results and the Lloyd-Max conditions, we isometrically embed  -- without loss of structural information -- graphs as points into some Riemannian orbifold. An orbifold is the quotient of a manifold by a finite group action and therefore generalizes the notion of manifold. Using orbifolds we can define geometric and analytic concept such as length, angle, derivative, gradient, and integral locally to a Euclidean space. This construction forms the basis for extending consistency results from Euclidean vector spaces to the domain of graphs.

The proposed approach has the following properties: First, it can be applied to finite combinatorial structures other than graphs like, for example, point patterns, sequences, trees, and hypergraphs. For the sake of concreteness, we restrict our attention exclusively to the domain of graphs. Second, for graphs consisting of a single vertex with  feature vectors as attributes, graph quantization coincides with vector quantization. Third, the proposed consistency results justify some of the above referenced graph clustering methods as statistically consistent learners. Fourth, the underlying mathematical framework can be applied in order to link other  structural pattern recognition methods that directly operate in the domain of graphs to methods from statistical pattern recognition.

The paper is organizes as follows. Section 2 describes the problem of graph quantizer design. Section 3 introduces Riemannian orbifolds. In Section 4, we extend VQ to GQ and present consistency result for GQ design techniques. Section 5 briefly discusses the case of general graph edit distance functions. Finally, Section 6 concludes.

 \section{The Problem of Graph Quantizer Design}

This section aims at outlining the problem of extending VQ to the quantization of graphs.

\subsection{Attributed Graphs}

To begin with, we first describe the structures we want to quantize. 

\introskip

Let $\S{A}$ be a set of \emph{attributes} and let $\varepsilon \in \S{A}$ be a distinguished element denoting the \emph{null} or \emph{void} element.  An  \emph{attributed graph} is a tuple $X = (V, \alpha)$ consisting of a finite nonempty set $V$ of \emph{vertices} and an \emph{attribute function} $\alpha: V \times V \rightarrow \S{A}$. Elements of the set
\[
E = \cbrace{(i,j)\in V \times V \,:\, i \neq j \text{ and }\alpha(i,j) \neq \varepsilon}
\]
are the \emph{edges} of $X$. By $\S{G_A}$ we denote the set of all attributed graphs with attributes from $\S{A}$. The vertex set of an attributed graph $X$ is often referred to as  $V_X$ and its attribute function as $\alpha_X$.  

An \emph{alignment} of a graph $X$ is a graph $X'$ with $V_X \subseteq V_{X'}$ and 
\[
\alpha_{X'}(i,j) = \begin{cases}
\alpha_X(i,j) & (i,j) \in V_X \times V_X\\
\varepsilon & \text{otherwise}
\end{cases}
\]
for all $i,j\in V_{X'}$. Thus, we obtain an alignment of $X$ by adding isolated vertices with null-attribute. The set $V_{X'}\setminus V_X$ is the set of \emph{aligned vertices}.
By $\S{A}(X)$ we denote the (infinite) set of all alignments of $X$.

A \emph{pairwise alignment} of graphs $X$ and $Y$ is a triple $(\phi, X',Y')$ consisting of  alignments $X'\in \S{A}(X)$ and $Y'\in \S{A}(Y)$ together with a bijective mapping
\[
\phi : V_{X'} \rightarrow V_{Y'}, \quad i \mapsto i^{\phi}.
\]
By $\S{A}(X,Y)$ we denote the set of all pairwise alignments between $X$ and $Y$. Sometimes we briefly write $\phi$ instead of $(\phi, X',Y')$.

\subsection{The Graph Edit Distance}

Fundamental for quantizing data is the notion of distortion. This section briefly introduces the graph edit distance functions as our choice of distortion measure. For a more detailed definition of the graph edit distance, we refer to \cite{Bunke94}. In addition, we present an important graph metric based on a generalization of the concept of  maximum common subgraph, which arises in various different guises as a common choice of proximity measure \cite{Almohamad93,Caetano07,Cour06,Gold96a,Umeyama88,Wyk02}. For sake of convenience, we assume that all distances are metrics.

\introskip

Each pairwise alignment $(\phi,X',Y') \in \S{A}(X,Y)$ can be regarded as an edit path with cost
\[
d_\phi\args{X,Y} = \sum_{i,j \in V_{X'}} d_{\S{A}}\args{\alpha_{X'}(i,j), \alpha_{Y'}(i^\phi, j^\phi)},
\]
where $d_\S{A}: \S{A} \times \S{A} \rightarrow \R_+$ is a distance function defined on the set $\S{A}$ of attributes. Observe that deletion (insertion) of vertices also deletes (inserts) all edges the respective vertices are incident to. 

The \emph{graph edit distance} of $X$ and $Y$ is then defined as the edit path with minimal cost 
\[
d(X, Y) = \min\cbrace{d_\phi\args{X,Y} \,:\, \phi \in \S{A}(X,Y)}.
\]
Note that the set $\S{A}(X,Y)$ of pairwise alignments is of infinite cardinality. But since $d_{\S{A}}(\varepsilon, \varepsilon) = 0$, we actually take the minimum over a finite subset by ignoring all pairwise alignments that map aligned vertices with null-attributes onto each other.

Next, we consider an important example of the graph edit distance based on a generalization of the concept of  maximum common subgraph. We derive this graph metric from a similarity measure in the same way the Euclidean distance is derived from an inner product.

Suppose that $k_{\S{A}}: \S{A} \times \S{A} \rightarrow \R$ with $k_{\S{A}}(\cdot, \varepsilon) = 0$ is a positive definite kernel. We measure the quality of a pairwise alignment $\phi\in \S{A}(X,Y)$ by
\[
k_\phi(X, Y) =  \sum_{i,j\in V_X} k_{\S{A}}\args{\alpha_X(i,j), \alpha_Y(i^\phi, j^\phi)}.
\]
An \emph{optimal alignment kernel} is a graph similarity measure of the form
\begin{align}\label{eq:oak}
k(X, Y) = \max\cbrace{k_\phi(X,Y)\,:\, \phi\in\S{A}(X, Y)}.
\end{align}
Note that $k\args{\cdot|\cdot}$ is symmetric but indefinite as a pointwise maximizer of a set of positive definite kernels. 

The distance metric on $\S{G_A}$ induced by an optimal alignment kernel $k\args{\cdot|\cdot}$ is defined by
\begin{align}\label{eq:intrinsic-mcs-metric}
d(X, Y) = \sqrt{l(X)^2 - 2k(X, Y) + l(Y)^2},
\end{align}
where $l(X) = \sqrt{k(X, X)}$ denotes the \emph{length} of an attributed graph $X$. As shown in \cite{Jain09c}, $d$ is indeed a metric and can be expressed as a graph edit distance.

\subsection{The Problem of Graph Quantizer Design}

Let $\args{\S{G_A}, d}$ be a graph distance space, where $d\args{\cdot|\cdot}$ is a graph edit distance. Optimal graph quantization design aims at minimizing the expected distortion
\[
D(\S{C}) = \int_{\S{G_A}} d\!\args{X, Q(X)} dP(X),
\]
where $Q:\S{G_A} \rightarrow \S{C}$ is a graph quantizer, $\S{C} = \cbrace{Y_1, \ldots, Y_k}$ a codebook consisting of $k$ code graphs, and $P = P_{\S{G_A}}$ is a probability measure defined on some appropriate measurable space $\args{\S{G_A}, \Sigma_{\S{G_A}}}$.

As opposed to vector quantization, the following factors complicate designing an optimal graph quantizer in a statistically consistent way: 
\begin{enumerate}
\item The graph distance $d(X, Y)$ is in general non-convex and non-differentiable. 
\item Neither a well-defined addition on graphs nor the notion of derivative for functions on graphs is known.
\end{enumerate}
To overcome these difficulties, we isometrically embed graphs as points into a Riemannian orbifold in order to apply methods that generalize gradient descent techniques and methods from stochastic optimization for non-convex and non-differentiable distortion functions.

\section{Riemannian Orbifolds}

Orbifolds generalize the notion of manifold as locally  being a quotient of $\R^n$ by finite group actions. Consequently, learning on orbifolds generalizes learning on Euclidean spaces and Riemannian manifolds. This section introduces Riemannian orbifolds and their intrinsic metric structure. Proofs for new results are delegated to Section \ref{subsec:proof:orbifold-functions}. For all other proofs we refer to \cite{Borzellino92,Jain09a}.

\subsection{Riemannian Orbifolds}

To keep the treatment simple, we assume that $\S{X} = \R^{n}$ is the $n$-dimensional Euclidean vector space, and $\Gamma$ is a permutation group acting on $\S{X}$. In a more general setting, however, we can assume that $\S{X}$ is a Riemannian manifold, and $\Gamma$ is a finite group of isometries acting effectively on $\S{X}$. 

The binary operation
\[
\cdot: \Gamma \times \S{X} \rightarrow \S{X}, \quad (\gamma, \vec{x}) \mapsto \gamma(\vec{x})
\]
is a group action of $\Gamma$ on $\S{X}$. For $\vec{x} \in \S{X}$, the \emph{orbit} of $\vec{x}$ is the set defined by
\[
\bracket{\vec{x}} = \cbrace{\gamma(\vec{x}) \,:\, \gamma \in \Gamma}.
\]	
The quotient set
\[
\S{X}_\Gamma = \S{X}/\Gamma = \cbrace{\bracket{\vec{x}} \,:\, \vec{x} \in \S{X}} 
\] 
consisting of all all orbits  carries the structure of a \emph{Riemannian orbifold}. Its \emph{orbifold chart} is the surjective continuous mapping 
\[
\pi: \S{X} \rightarrow \S{X}_\Gamma, \quad \vec{x} \mapsto \bracket{\vec{x}}
\]
that projects each point $\vec{x}$ to its orbit $\bracket{\vec{x}}$.

In the following, an orbifold is a triple $\S{Q} = \args{\S{X}, \Gamma, \pi}$ consisting of an Euclidean space $\S{X}$, a permutation group $\Gamma$ acting on $\S{X}$ and its orbifold chart $\pi$. With $\Gamma =\cbrace{\id}$ being the trivial permutation group consisting of the identity only, a manifold $\S{X}$ is also an orbifold. In general, however, the underlying space $\S{X}_\Gamma$ of an orbifold is not a manifold. Thus, orbifolds generalize the notion of manifold. The points at which an orbifold $\S{X}_\Gamma$ is locally not homeomorphic to a manifold are its \emph{singular points}. We call the elements of $\S{X}_\Gamma$ \emph{structures}, since they represent combinatorial structures like attributed graphs. We use capital letters $X, Y, Z, \ldots$ to denote structures from $\S{X}_\Gamma$ and write, by abuse of notation, $\vec{x} \in X$ if $\pi(\vec{x}) = X$. Each vector $\vec{x} \in X$ is a \emph{vector representation} of structure $X$ and the set $\S{X}$  of all vector representation is the \emph{representation space} of $\S{X}_\Gamma$.

\begin{example}\label{ex:simple1}
 Let $\S{X} = \R^2$ and let $\Gamma$ be the group generated  by reflections across the main-diagonal of the x-y-plane. Then $\S{Q} = \args{\S{X}_\Gamma, \Gamma, \pi}$ is a Riemannian orbifold with 
\[
\pi :\S{X} \rightarrow \S{X}_\Gamma, \quad \vec{x} = (x_1,x_2) \mapsto \bracket{\vec{x}} = \cbrace{(x_1,x_2), (x_2, x_1)}.
\]
The singular points of $\S{X}_\Gamma$ are all structures $X$ represented by vectors $\vec{x} = (x_1, x_2)$ with $x_1 = x_2$.
\end{example}

\subsection{The Riemannian Orbifold of Attributed Graphs}

In this section, we show that attributes graphs can be identified with points in some Riemannian orbifold. 

\introskip

Riemannian orbifolds of attributed graphs arise by considering equivalence classes of matrices representing the same graph. To identify graphs with points in a Riemannian orbifold without loss of structural information, some technical assumptions  and restrictions to simplify the mathematical treatment are necessary. For this, let $\args{\S{G_A}, d}$ be a graph distance space with graph edit distance $d(\cdot|\cdot)$. Then we make the following assumptions:
\begin{description}
\item[P1] There is a feature map $\Phi:\S{A} \rightarrow \S{H}$ of the attributes into some finite dimensional Euclidean feature space $\S{H}$ and a distance function $d_{\S{H}}: \S{H}\times \S{H} \rightarrow \R_+$ such that $\Phi(\varepsilon) = \vec{0} \in \S{H}$ and
\[
d_{\S{A}}(a,a') = d_{\S{H}}(\Phi(a), \Phi(a'))
\]
for all attributes $a, a'\in \S{A}$.
\item[P2] All graphs are finite of bounded order $n$, where $n$ is a sufficiently large number. Graphs $X$ of order less than $n$, say $m < n$, are aligned to graphs $X'$ of order $n$ by inserting $p = n-m$ isolated vertices with null attribute $\varepsilon$. 
\end{description}

Before discussing the impact of both assumptions for practical application, we first restate our first assumptions for graph metrics induced by optimal alignment kernels. By definition $k_{\S{A}}: \S{A} \times \S{A} \rightarrow \R$ is a positive definite kernel corresponding to an inner product $k_{\S{A}}(x, y) = \inner{\Phi(x), \Phi(y)}$ in some feature space $\S{H}$. 
Our first assumption requires that $\S{H}$ is a finite dimensional Euclidean space and $\Phi(\varepsilon) = \vec{0}$.

Now let us consider the above assumptions in more detail. Both conditions do not effect the graph edit distance, provided an appropriate feature map for the attributes can be found. Restricting to finite dimensional Euclidean feature spaces $\S{H}$ is necessary for deriving consistency results and for applying methods from stochastic optimization. Limiting the maximum size of the graphs to some arbitrarily large number $n$ and aligning smaller graphs to graphs of oder $n$ are purely technical assumptions to simplify mathematics. For machine learning problems, this limitation should have no practical impact, because neither the bound $n$ needs to be specified explicitly nor an extension of all graphs to an identical order needs to be performed. When applying the theory, all we actually require is that the order of the graphs is bounded.

With both assumptions in mind, we construct the Riemannian orbifold of attributed graphs. Let $\S{X} = \S{H}^{n \times n}$ be the set of all $(n \times n)$-matrices with elements from feature space $\S{H}$. A graph $X$ is completely specified by a \emph{representation matrix} $\vec{X} = \args{\vec{x}_{ij}}$ from $\S{X}$ with elements 
\[
\vec{x}_{ij} = \begin{cases}
\phi\args{\mu_X(i)} & i = j \\
\phi\args{\nu_X(i,j)} & (i,j) \in E \\
\vec{0} & \mbox{otherwise}
\end{cases}
\]
for all $i, j \in V_X$. The form of a representation matrix $\vec{X}$ of $X$ is generally not unique and depends on how the vertices are arranged in the diagonal of $\vec{X}$.

Now suppose that $\Pi^n$ be the set of all $(n\times n)$-permutation matrices. For each $\vec{P} \in \Pi^n$ we define a mapping
\[
\gamma_{\vec{P}} : \S{X} \rightarrow \S{X}, \quad \vec{X} \mapsto \vec{P}^{\T} \vec{X}\vec{P}.
\]
Then $\Gamma = \cbrace{\gamma_{\vec{P}}\,:\, \vec{P} \in \Pi^n}$ is a permutation group acting on $\S{X}$. Regarding an arbitrary matrix $\vec{X}$ as a representation of some graph $X$,  then the orbit $\bracket{\vec{X}}$ consists of all possible matrices that can represent $X$. By identifying the orbits of $\S{X}_\Gamma$ with attributed graphs,  the set $\S{G_A}$ of attributed graphs of bounded order $n$ is a Riemannian orbifold.

\subsection{Metric Structures}
Let  $\S{Q} = \args{\S{X}, \Gamma, \pi}$ be an orbifold. We derive an intrinsic  metric that enables us to do Riemannian geometry. In the case of a Riemannian orbifold of attributed graphs the intrinsic metric coincides with the graph metric of (\ref{eq:intrinsic-mcs-metric}) induced by an optimal alignment kernel.

\introskip

Any inner product $\inner{\cdot, \cdot}$ on $\S{X}$ gives rise to a maximizer of the form
\[
k: \S{X}_\Gamma \times \S{X}_\Gamma \rightarrow \R, \quad \args{X, Y} \mapsto \max \cbrace{\inner{\vec{x}, \vec{y}} \,:\, \vec{x} \in X, \vec{y} \in Y}. 
\]
We call the kernel function $k(\cdot | \cdot)$ \emph{optimal alignment kernel}, induced by the inner product $\inner{\cdot, \cdot}$. Note that the maximizer of a set of positive definite kernels is an indefinite kernel in general. Since $\Gamma$ is a group, we find that 
\[
k(X, Y) = \max \cbrace{\inner{\vec{x}, \vec{y}} \,:\, \vec{x} \in X}.
\]
where $\vec{y}$ is an arbitrary but fixed vector representation of $Y$. In general, we have 
\[
k(X, Y) \geq \inner{\vec{x}, \vec{y}}
\] 
for all $\vec{x}\in X$ and $\vec{y}\in Y$.  

\begin{example}\label{ex:simple2}
Consider the Riemannian orbifold $(\S{X}, \Gamma, \pi)$ of Example \ref{ex:simple1}, where $\S{X} = \R^2$ and $\Gamma = \cbrace{\id, \gamma}$ is the group generated by reflections across the x-y-plane. Suppose that $\vec{x} = (1, 2)$ is a vector representation of $X$ and $\vec{y} = (3, 2)$ is a vector representation of $Y$. Then the optimal alignment kernel $k\args{X, Y}$ induced by the standard inner product of $\S{X}$ is given by 
\begin{align*}
k(X, Y) &= \max\cbrace{\inner{\vec{x}, \vec{y}}, \inner{\gamma(\vec{x}), \vec{y}}, \inner{\vec{x}, \gamma(\vec{y}}),  \inner{\gamma(\vec{x}), \gamma(\vec{y}})}
\end{align*}
Evaluating the inner products yields
\begin{align*}
\inner{\vec{x}, \vec{y}} &= \inner{(1,2), (3,2)} = 7\\
\inner{\gamma(\vec{x}), \vec{y}} &= \inner{(2,1), (3,2)} = 8\\
\inner{\vec{x}, \gamma(\vec{y})} &= \inner{(1,2), (2,3)} = 8\\
\inner{\gamma(\vec{x}), \gamma(\vec{y})} &= \inner{(2,1), (2,3)} = 7.
\end{align*}
Thus, we have $k(X,Y) = 8$. 
\end{example}

\begin{example}
Suppose that $X$ and $Y$ are attributed graphs where edges have attribute $1$ and vertices have attribute $0$. The optimal alignment kernel $k\args{X, Y}$ induced by the standard inner product of $\S{X}$ is the number of edges of a maximum common subgraph of $X$ and $Y$. 
\end{example}

\begin{example}
More generally, if property P1 is satisfied, then any optimal alignment kernel on a bounded set of attributed graphs  as defined in (\ref{eq:oak}) is also an optimal assignment kernel of some Riemannian orbifold.
\end{example}

Suppose that $X \in \S{X}_\Gamma$. Since $k(X, X) = \inner{\vec{x}, \vec{x}}$ for all $\vec{x} \in X$, we can define the \emph{length} of $X$  by 
\[
l(X) = \sqrt{k(X, X)}.
\]
The optimal alignment kernel together with the length satisfies the Cauchy-Schwarz inequality
\[
\abs{k(X, Y)} \leq l(X)\cdot l(Y).
\]
Since the Cauchy-Schwarz inequality is valid, the geometric interpretation of $k(\cdot|\cdot)$ is that it computes the cosine of a well-defined angle between $X$ and $X'$ provided they are normalized to length $1$.

Likewise, $k(\cdot|\cdot)$ gives rise to a distance function defined by
\[
d(X, Y) = \sqrt{l(X)^2 -2 k(X, Y) +\, l(Y)}.
\]
From the definition of $k(\cdot|\cdot)$ follows that $d$ is a metric. In addition, we have
\begin{align}\label{eq:mcs-metric-2}
d(X, Y) =  \min \cbrace{\norm{\vec{x} - \vec{y}} \,:\, \vec{x} \in X, \vec{y} \in Y}, 
\end{align}
where $\norm{\cdot}$ denotes the Euclidean norm induced by the inner product $\inner{\cdot, \cdot}$ of the Euclidean space $\S{X}$. 

\begin{example}\label{ex:simple3}
Consider the Riemannian orbifold $(\S{X}, \Gamma, \pi)$ of Example \ref{ex:simple1} and \ref{ex:simple2}. Suppose that $\vec{x} = (1, 2)$ is a vector representation of $X$ and $\vec{y} = (3, 2)$ is a vector representation of $Y$. Then the squared lengths of $X$ and $Y$ are $l(X)^2 = 5$ and $l(Y)^2= 13$. Since $k(X, Y) = 8$ according to Example \ref{ex:simple2}, the distance is $d(X,Y) = \sqrt{5 - 16 + 13} = \sqrt{2}$.
\end{example}

\begin{example}
If properties P1 and P2 are satisfied, then the graph metric (\ref{eq:intrinsic-mcs-metric}) coincides with the intrinsic orbifold metric (\ref{eq:mcs-metric-2}).
\end{example}

Equation (\ref{eq:mcs-metric-2}) states that $d\args{\cdot|\cdot}$ is the length of a minimizing geodesic of $X$ and $Y$ and therefore an intrinsic metric, because it coincides with the infimum of the length of all admissible curves from $X$ to $Y$.   In addition, we find that the topology of $\S{X}_\Gamma$ induced by the metric $d$ coincides with the quotient topology induced by the topology of the Euclidean space $\S{X}$. 

\subsection{Orbifold Functions}

Suppose that $\S{Q} = \args{\S{X}, \Gamma, \pi}$ is an orbifold. An \emph{orbifold function} is a mapping
\[
f:\S{X}_{\Gamma} \rightarrow \R.
\]
The \emph{lift} of $f$ is a function 
\[
\tilde{f}: \S{X} \rightarrow \R
\]
satisfying $\tilde{f} = f \circ \pi$. The lift $\tilde{f}$ is invariant under group actions of $\Gamma$, that is $\tilde{f}(\vec{x}) = \tilde{f}\args{\gamma(\vec{x})}$ for all $\gamma \in \Gamma$.

We say, an orbifold function $f:\S{X}_{\Gamma} \rightarrow \R$ is continuous (locally Lipschitz, differentiable, generalized differentiable) at $X \in \S{X}_\Gamma$ if its lift $\tilde{f}$ is continuous (locally Lipschitz, differentiable, generalized differentiable) at some vector representation $\vec{x} \in X$. The definition is independent of the choice of the vector representation that projects to $X$ (see Section \ref{subsec:proof:orbifold-functions}, Prop.\ \ref{prop:con} -- Prop.\ \ref{prop:gendiff}). For a  definition of generalized differentiable functions and their basic properties we refer to Section \ref{sec:GDF}.

\begin{example}
Consider the Riemannian orbifold $(\S{X}, \Gamma, \pi)$ of Example \ref{ex:simple1}-\ref{ex:simple3}. The function
\[
f_Y:\S{X}_\Gamma \rightarrow \R, \quad X \mapsto k(X,Y)
\]
for some $Y\in \S{X}_\Gamma$ is an orbifold function with lift
\[
\tilde{f}_Y:\S{X} \rightarrow \R, \quad \vec{x} \mapsto \max \cbrace{\inner{\vec{x}, \vec{y}}, \inner{\vec{x}, \gamma(\vec{y})}},
\]
 where $\vec{y} \in Y$. Analytical properties of $f$ such as continuity and differentiability can be investigated using the lift $\tilde{f}$ of $f$. For example, if $\tilde{f}$ is differentiable at $\vec{x} \in X$ then it is also differentiable at $\gamma(\vec{x})$ according to Prop.\ \ref{prop:diff}. Hence, differentiability of the orbifold function $f$ is well-defined at $X$. 
\end{example}

\subsection{Gradients and Generalized Gradients of Orbifold Functions}

We extend the notion of gradient and generalized gradient to differentiable and generalized differentiable orbifold functions. 
\paragraph*{Gradient of Differentiable Orbifold Functions.}
Suppose that $f: \S{X}_{\Gamma} \rightarrow \R$ is differentiable at $X \in \S{X}_\Gamma$. Then its lift $\tilde{f}:\S{X} \rightarrow \R$ is differentiable at all vector representations that project to $X$. The \emph{gradient} $\nabla f(X)$ of $f$ at $X$ is defined by the projection
\[
\nabla f(X) = \pi\args{\nabla \tilde{f}(\vec{x})} 
\]
of the gradient $\nabla \tilde{f}(\vec{x})$ of $\tilde{f}$ at a vector representation $\vec{x} \in X$. This definition is independent of the choice of the vector representation. We have
\[
\nabla \tilde{f}(\gamma(\vec{x})) = \gamma\args{\nabla\tilde{f}(\vec{x})}
\]
for all $\gamma \in \Gamma$. This implies that the gradients of $\tilde{f}$ at $\vec{x}$ and $\gamma(\vec{x})$ are vector representations of the same structure, namely the gradient $\nabla f(X)$ of the orbifold function $f$ at $X$. Thus, the gradient of $f$ at $X$ is a well-defined structure pointing to the direction of steepest ascent  (see Section \ref{subsec:proof:orbifold-functions}, Prop.\ \ref{prop:diff}).

\paragraph*{Subdifferential of Generalized Differentiable Orbifold Functions.}
Suppose that $f: \S{X}_{\Gamma} \rightarrow \R$ is generalized differentiable at $X \in \S{X}_\Gamma$. Then its lift $\tilde{f}:\S{X} \rightarrow \R$ is generalized differentiable at all vector representations that project to $X$. The \emph{subdifferential} $\partial f(X)$ of $f$ at $X$ is defined by the projection
\[
\partial f(X) = \pi\args{\partial \tilde{f}(\vec{x})} 
\]
of the subdifferential $\partial \tilde{f}(\vec{x})$ of $\tilde{f}$ at a vector representation $\vec{x} \in X$. This definition is independent of the choice of the vector representation. We have 
\[
\partial \tilde{f}(\gamma(\vec{x})) = \gamma\args{\partial\tilde{f}(\vec{x})}
\]
for all $\gamma \in \Gamma$. This implies that the subdifferentials $\partial\tilde{f}(\vec{x}) \subseteq \S{X}$ and $\partial\tilde{f}(\gamma(\vec{x})) \subseteq \S{X}$ are subsets that project to the same subset of $\S{X}_\Gamma$, namely the subdifferential $\partial f(X)$ (see Section \ref{subsec:proof:orbifold-functions}, Prop.\ \ref{prop:gendiff}).

The properties of generalized differentiable function as listed in Section \ref{sec:GDF} carry over to generalized differentiable orbifold functions via their lifts. For example, a generalized differentiable orbifold function is locally Lipschitz and therefore differentiable almost everywhere.

\begin{example}\label{ex:gd-distortion1}
Let $(\S{G_A}, d)$ be a graph space, where 
\[
d(X,Y) = \min_{\phi\in\S{A}(X,Y)} d_\phi(X,Y)
\]
is a graph edit distance. We can identify $\S{G_A}$ with a Riemannian orbifold $\S{Q} = (\S{X}, \Gamma, \pi)$ and the graph edit distance $d\args{\cdot|\cdot}$ with a distance function defined on $\S{X}_\Gamma$. Suppose that the cost functions $d_\phi\args{\cdot|\cdot}$ of the edit paths are continuously differentiable (generalized differentiable). Then the distance  $d\args{\cdot|\cdot}$ is generalized differentiable.
\end{example}

\begin{example}\label{ex:gd-distortion2}
Let $\S{Q}$ be a Riemannian orbifold of attributed graphs. Then (i) an optimal assignment kernel $k\args{\cdot|\cdot}$, (ii) the intrinsic metric  $d\args{\cdot|\cdot}$ induced by $k\args{\cdot|\cdot}$, and (iii) the squared metric $d\args{\cdot|\cdot}^{2}$ are generalized differentiable. 
\end{example}

\subsection{Integration on Orbifolds}

Suppose that $\S{Q}= \args{\S{X}, \Gamma, \pi}$ is a Riemannian orbifold with singular set $\S{S_Q}$. In order to integrate orbifold functions $f:\S{X}_{\Gamma} \rightarrow \R$ by the Lebesgue integral, we need to construct an appropriate measurable space together with an orbifold measure. The measurable space is defined by the Borel set $\S{B}(\S{X}_\Gamma)$ generated by the open sets of $\S{X}_\Gamma$.  From the orbifold measure we expect that it is compatible with the local Riemannian measures. In addition, we demand that the singular set $\S{S_Q}$ has measure $0$. This is motivated by the following fact: The singular set is covered locally by the finite union of totally geodesic submanifolds, which has measure $0$ relative to the local canonical Riemannian measure. Since the projection to the orbifold is distance decreasing, it is reasonable to ask for an orbifold measure that assigns  measure $0$ to the singular set $\S{S_Q}$.

Let  $\S{B}\args{\S{X}_\Gamma \setminus \S{S_Q}}$ denote the Borel set generated by the open sets of $\S{X}_\Gamma \setminus \S{S_Q}$. Then there exists a complete canonical measure $\mu$ on the the Borel set $\S{B}\args{\S{X}_\Gamma \setminus \S{S_Q}}$ given by a unique volume form on $\S{X}_\Gamma \setminus \S{S_Q}$. The measure $\mu$ can be extended to a complete measure $\nu$ on the Borel set 
$\S{B}(\S{X}_{\Gamma})$ such that
\[
\nu\args{\S{A}} = \mu\args{\S{A}\setminus\S{S_Q}} = \int_{\S{A}\setminus\S{S_Q}} d\mu.
\]
In particular, we have $\nu(\S{A}) = 0$ for any subset $\S{A}\subseteq \S{S_Q}$. For proofs we refer to \cite{Borzellino92}.

In the following we write 
\[
\int_{\S{U}_\Gamma} f(X)dX = \int_{\S{U}_\Gamma} f d\nu
\]
for the integral of an orbifold function $f:\S{U}_\Gamma \rightarrow \R$ defined on a measurable subset $\S{U}_\Gamma \subseteq \S{X}_\Gamma$. We tacitly assume that all integrals occurring in the following sections exist.

\section{Graph Quantization}

This section extends vector quantization to quantization of graphs. 

\subsection{The Basics}

Suppose that $\S{Q} = \args{\S{X}, \Gamma, \pi}$ is a Riemannian orbifold. A \emph{graph quantizer} of size $k$ is a mapping of the form
\[
Q : \S{X}_\Gamma \rightarrow \S{C}
\]
where $\S{C} = \cbrace{Y_1, \ldots, Y_k} \subseteq \S{X}_\Gamma$ is a finite set, called \emph{codebook}. The elements $Y_j\in\S{C}$ are the \emph{code graphs}. The graph quantizer $Q$ partitions the input space $\S{X}_\Gamma$ into $k$ disjoint \emph{regions}
\[
\S{R}_j = \cbrace{X \in \S{X}_\Gamma \,:\, Q(X) = Y_j}
\]
such that their union covers $\S{X}_\Gamma$. By $\S{P}_Q$ we denote the partition of $Q$ consisting of all $k$ regions $\S{R}_j$.

Suppose that $\S{J} = \cbrace{1, \ldots, k}$. The basic operation of a vector quantizer $Q$ can be written as a composition $Q = d_Q \circ e_Q$ of an \emph{encoder} 
$e_Q:\S{X}_\Gamma \rightarrow \S{J}$ and a \emph{decoder} $d_Q:\S{J} \rightarrow \S{C}$. The encoder assigns each input graph to a region via the index set $\S{J}$. The decoder maps indices of $\S{J}$ referring to regions to code graphs.

\subsection{Graph Quantizer Performance}
We measure the performance of a graph quantizer  $Q$ by the expected distortion
\[
D(Q)= \E_{X} \bracket{d\args{X, Q(X)}} = \int_{\S{X}_\Gamma} d(X, Q(X)) dP(X),
\]
where $X \in \S{X}_\Gamma$ is a random variable with probability measure $P = P_{\S{X}_\Gamma}$ representing the observable graphs to be quantized. The expectation $\E_X$ is taken with respect to some probability space $\args{\S{X}_\Gamma, \Sigma_{\S{X}_\Gamma}, P_{\S{X}_\Gamma}}$. The quantity $d(X, Y)$ measures the \emph{distortion} of the random input graph $X$ and code graph $Y$. Here we consider graph distortion measures that are graph edit distances. An example is the squared metric induced by an optimal alignment kernel
\[
d\args{X, Y} = \min_{\vec{x} \in X, \vec{y}\in Y}\normS{\vec{x}-\vec{y}}{^2}
\]
Using the codebook and partition for the given quantizer $Q$, we can rewrite the expected distortion by
\[
D(\S{C}) = \sum_{j=1}^{k} \int_{\S{R}_j} d(X, Y) dP(X).
\]

\subsection{The Problem of Optimal Graph Quantizer Design}

The problem of optimal graph quantizer design is stated as follows: Find a codebook $\S{C}$ specifying the decoder $d_Q$ and a partition $\S{P}_Q$ specifying the encoder $e_Q$ such that the expected distortion $D(Q)$ is minimized. The composite mapping  $Q = d_Q \circ e_Q$ of the resulting encoder and decoder is then an \emph{optimal graph quantizer}. 

An optimal graph quantizer satisfies the following necessary conditions, also known as the \emph{Lloyd-Max conditions}:
\begin{enumerate}
\item \emph{Nearest Neighbor Condition}. Given a fixed codebook $\S{C}$, a graph quantizer $Q$ is optimal, if the code vector $Q(X)$ of an input pattern $X$ satisfies the nearest neighbor rule
\[
Q(X) = \arg\min_{Y \in \S{C}}  d\args{X, Y}
\]
for all $X\in \S{X}_\Gamma$, where ties are resolved according to some rule. A proof is given in Section \ref{subsec:NCGQ}, Theorem \ref{theorem:NNC}. 
\item \emph{Centroid Condition}. Given a fixed partition $\S{P}_Q$, a vector quantizer $Q$ is optimal, if each code vector $Y_j$ is the centroid of region $\S{R}_j$, that is
\[
Y_j = \arg\min_{Y\in \S{X}_\Gamma} \E\bracket{d\args{X, Y}\,|\, X\in\S{R}_j}
\]
for all $Y\in \S{X}_\Gamma$ and all $j \in \S{J}$. A proof is given in Section \ref{subsec:NCGQ}, Theorem \ref{theorem:CC}.
\end{enumerate}
Note that $Y_j$ with 
\[
Y_j = \arg\min_{Y\in \S{X}_\Gamma} \E\bracket{d\args{X, Y}\,|\, X\in\S{R}_j}
\]
is called a \emph{centroid} of region $\S{R}_j$. The centroids may not be unique. This also holds for squared metrics induced by some optimal assignment kernel, which are the counterparts of squared Euclidean distances.

\subsection{Graph Quantizer Design}

Since the distribution $P = P_{\S{X}_\Gamma}$ of the observable graphs is usually unknown, the expected distortion $D(\S{C})$ can neither be computed nor be minimized directly. Instead, we design (estimate) an optimal quantizer from empirical data. For vectors, prominent methods for designing an optimal quantizer are k-means and simple competitive learning. Both methods, k-means and simple competitive learning have been extended for designing graph quantizers in the context of prototype based clustering. To derive consistency results for k-means and simple competitive learning in the domain of graphs, we consider estimators based on empirical distortions and on stochastic approximation.

\subsubsection{Estimators based on Empirical Distortion Measures.} In order to derive consistency results, we restrict the set of feasible codebooks to a compact subspace
\[
\S{W} \subset  \S{X}_\Gamma^k = \underbrace{\S{X}_\Gamma \times \cdots \times \S{X}_\Gamma}_{\text{k-times}} 
\]
of the topological space $\S{X}_\Gamma^k$. The problem of designing an optimal quantizer for graphs is then of the form
\[
\min_{\S{C} \in \S{W}}\quad D(\S{C}) = \sum_{j=1}^{k} \int_{\S{R}_j} d(X, Y) dP(X).
\]
where the minimum is taken over the compact set $\S{W}$ rather than $\S{X}_\Gamma^k$.  Let 
\begin{enumerate}
\item  $D^*$ be the set of minimal values of the expected distortion $D(\S{C})$,
\item $\S{W}^*  = \cbrace{\S{C} \in \S{W} \,:\, D(\S{C}) = D^*} $ be the set of true (optimal) codebooks, and
\item $\S{W}^*_{\varepsilon} = \cbrace{\S{C} \in \S{W} \,:\, D(\S{C}) \leq D^* + \varepsilon}$ be the set of approximate solutions.
\end{enumerate}

To design an optimal graph quantizer, we minimize the \emph{empirical distortion}
\[
\hat{D}_N(\S{C})= \frac{1}{N}\sum_{i=1}^N  \min_{j \in \S{J}} d\argsS{X_i,Y_j},
\]
where $\S{C} \in \S{W}$ and $\S{S}= \cbrace{X_{1}, \ldots, X_{N}}$ is a training set consisting of $N$ independent graphs $X_i$ drawn from $\S{X}_\Gamma$. Let
\begin{enumerate}
\item  $\hat{D}_N^*$ be the set of minimal values of the empirical distortion $\hat{D}_N(\S{C})$,
\item $\S{W}_N^*  =  \{\S{C} \in \S{W} \,:\, \hat{D}_N(\S{C}) = \hat{D}^*_N\}$ be the set of empirical codebooks, and
\item $\S{W}_{N\varepsilon}^* = \{\S{C} \in \S{W} \,:\, \hat{D}_N(\S{C}) \leq \hat{D}^*_N + \varepsilon\}$ be the set of approximate solutions.
\end{enumerate}
The next result shows that estimators based on empirical distortions are consistent estimators.   
\begin{theorem}\label{theorem:consistency1}
Suppose that $\S{Q} = \args{\S{X}, \Gamma, \pi}$ is a Riemannian orbifold,  $d(X,Y)$ is a locally Lipschitz metric on $\S{X}_\Gamma$ with integrable Lipschitz constant, and $\S{W} \subseteq \S{X}_\Gamma^k$ is compact.
Then we have
\begin{align*}
\lim_{N \to \infty} \hat{D}_N^*\args{\omega} = D^*\\
\lim_{N \to \infty} \S{W}_N^*\args{\omega} = \S{W}^*\\
\lim_{N \to \infty} \S{W}_{\epsilon N}^*\args{\omega} = \S{W}_{\epsilon}^*
\end{align*}
almost surely.
\end{theorem}
The proof follows from \cite{Ermoliev91} applied to the lift $\tilde{d}$ of distortion $d$. Examples of locally Lipschitz distance metrics on $\S{X}_\Gamma$ with integrable Lipschitz constants are metrics induced by an optimal alignment kernel
\[
d(X, Y) = \min_{\vec{x}\in X, \vec{y}\in Y} \norm{\vec{x}-\vec{y}} 
\]
as well as $d(X,Y)^2$. 

\paragraph*{K-Means.}
In order to extend the standard k-means method to graphs for constructing an empirical codebook, we use the following update rule
\[
\vec{y}_{j}^{t+1} = \frac{1}{N_j^t} \sum_{i=1}^{N} q_{ij}^t\vec{x}_i,
\]
where $t > 0$ is the iteration, $\vec{x}_i \in X_i$ and $\vec{y}_j^t \in Y_j^t$ are vector representations that are optimally aligned,\footnote{Recall that two vector representations $\vec{x} \in X$ and $\vec{y}\in Y$ are optimally aligned if $\norm{\vec{x} - \vec{y}} = d(X, Y)$} and $\vec{Q}^t =\args{q_{ij}^t}$ is the matrix representation of the nearest neighbor quantizer $Q^t$ restricted to the training set $\S{S}$. The elements of $\vec{Q}^t$ are of the form
\[
q_{ij}^t = \begin{cases}
1 & Q^t(X_i) = Y_j^t\\
0 & \text{otherwise}
\end{cases}.
\]
The quantity $N_j^t$ denotes the number of elements from the training sets that are quantized by code graph  $Y_j^t$.

As for vectors, a drawback of k-means for graphs is that it is a local optimization technique for which existing consistency theorems are inapplicable, because Theorem \ref{theorem:consistency1} assumes global instead of local minimizers of the empirical distortion as estimators.

\subsubsection{Estimators based on Stochastic Optimization.}
Suppose that $\S{W} = \S{X}_\Gamma^k$. Stochastic optimization methods directly minimize the expected distortion
\begin{align*}
D\args{\S{C}} &= \sum_{j=1}^{k} \int_{\S{R}_j} d \argsS{X, Y_j} dP(X)\\
&=  \sum_{j=1}^{k} \int_{\S{X}_\Gamma} \min_{1 \leq j \leq k} d\argsS{X, Y_j} dP(X),
\end{align*}
using a training set $\S{S}= \cbrace{X_{1}, \ldots, X_{N}}$ of $N$ independent graphs $X_i$ drawn from $\S{X}_\Gamma$.  We assume that the loss function
\[
L(X, \S{C}) =  \min_{1 \leq j \leq k} d\args{X, Y_j}
\]
is generalized-differentiable, hence $L(X, \S{C})$ is differentiable almost everywhere. 

\begin{example}
If he graph distortion $d(\cdot|\cdot)$ is generalized differentiable, then the loss function $L(X, \S{C})$ is also generalized differentiable by calculus of generalized differentiable functions. This holds for graph distortions of Example \ref{ex:gd-distortion1} and \ref{ex:gd-distortion2}.
\end{example}

Since  the interchange of integral and generalized gradient remains valid for generalized differentiable loss functions, that is 
\[
\partial D(\S{C}) = \E_X\bracket{\partial  L(X,\S{C})}
\]
under mild assumptions (see \cite{Ermoliev1998,Norkin1986}), we can minimize the expected distortion $D(\S{C})$ according to the following \emph{stochastic generalized gradient} (SGG) method:
\begin{align}\label{eq:sggm}
\vec{y}_{t+1} &= \vec{y}_t + \eta_t  \args{\vec{x}_t - \vec{y}_t},
\end{align}
where $\vec{x}_t$ is a vector representation of input pattern $X_t \in \S{S}$, which is optimally aligned to vector representation $\vec{y}_t$ of a code graph $Y_t$ closest to $X_t$.
The random elements $\vec{s}_t = \vec{x}_t - \vec{y}_t \in S_t$ are vector representations of \emph{stochastic generalized gradients} $S_t$, i.e.\ random variables defined on the probability space $\argsS{\S{X}_\Gamma, \Sigma_{\S{X}_\Gamma}, P_{\S{X}_\Gamma}}{^\infty}$ such that 
\begin{align}\label{eq:A2}
\E\bracket{S_t \,|\, \S{C}_0, \ldots, \S{C}_t} \in \partial D\args{\S{C}}.
\end{align}
We consider the following conditions for almost sure convergence of stochastic optimization:
\begin{description}
\item[A1] The sequence $(\eta_t)_{t \geq 0}$ of step sizes satisfies 
\[
\eta_t > 0, \quad \lim_{t \to \infty} \eta_t = 0, \quad \sum_{t=1}^{\infty} \eta_t = \infty, \quad \sum_{t=1}^{\infty} \eta_t^2 < \infty.
\]
\item[A2] The stochastic generalized gradients  $\args{S_t}_{t \geq 0}$ satisfy (\ref{eq:A2}).
\item[A3] The expected squared norm of stochastic generalized gradients $\args{S_t}_{t \geq 0}$ is bounded by 
\[
\E\bracket{\normS{S_t}{^2}} < +\infty.
\]
\end{description}

The next result shows that the SGG method is a consistent estimator.   
\begin{theorem}\label{theorem:consistency2}
Let $\S{Q} = \args{\S{X}, \Gamma, \pi}$ be a Riemannian orbifold and let $d(X,Y)$ be a generalized differentiable metric on $\S{X}_\Gamma$. Suppose that assumptions $(A1)-(A3)$ hold. Then the sequence $\argsS{\S{C}_t}{_{t\geq 0}}$ generated by the SGG method converges almost surely to graphs  satisfying necessary extremum conditions 
\[
\S{W}^* = \cbrace{\S{C} \in\S{W} \,:\; 0 \in \partial D(\S{C}) }.
\]
Besides the sequence $\argsS{D(\S{C}_t)}{_{t\geq 0}}$ converges almost surely and we have
\[
\lim_{t\to \infty} D(\S{C}_t) \in D(\S{W}^*).
\]
\end{theorem}
The proof is a direct consequence of Ermoliev and Norkin's Theorem \cite{Ermoliev1998} applied on the lift $\tilde{d}\args{\cdot|\cdot}$ of $d\args{\cdot|\cdot}$. 

\section{Remarks to GQ using the Graph Edit Distance}

In many applications, the graph edit distance is discontinuous. Examples include edit distances with constant non-zero deletion and/or insertion cost. A necessary (but not sufficient) condition for the consistency results stated in Theorem \ref{theorem:consistency1}  and \ref{theorem:consistency2} is that the underlying graph distortion is locally Lipschitz. Hence, both consistency results are inapplicable for discontinuous graph distortions. Let us consider both cases separately.

\paragraph*{Estimators based on Empirical Distortion Measures.}
Estimators based on empirical distortion measures aim at approximating the expected distortion $D(\S{C})$ by its empirical mean
\[
\min_{\S{C}\in \S{W}} \quad \hat{D}_N(\S{C})= \frac{1}{N}\sum_{i=1}^N  \min_{j \in \S{J}} d\argsS{X_i,Y_j}.
\]
As shown in \cite{Ermoliev97}, minimizing the empirical distortion is often meaningless, if the underlying graph edit distance function $d\args{\cdot|\cdot}$ and thus $\hat{D}_N(\S{C})$ is discontinuous, even if  the expectation $D(\S{C})$ may be continuously differentiable. Since the local solutions of $\hat{D}_N(\S{C})$ may have nothing in common with the local solutions of the original problem, estimators based on the empirical distortion $\hat{D}_N(\S{C})$ can be statistically inconsistent. Hence, minimizing $\hat{D}_N(\S{C})$ with underlying discontinuous graph edit distance using global or local optimization techniques like, for example, k-means lacks theoretical support. 

\paragraph*{Estimators based on Stochastic Optimization.}
The situation is better for estimators based on methods from stochastic optimization. For discontinuous graph edit distances $d\args{\cdot|\cdot}$ the expected distortion can be minimized in a statistically consistent way, for example, by methods based on approximations of $d\args{\cdot|\cdot}$ via averaged functions obtained by convolution with so-called mollifiers. For details, we refer to \cite{Ermoliev95}.

\section{Conclusion}
This contribution proposes a theoretical sound foundation of graph quantization generalizing the ideas of vector quantizations to the domain of attributed graph. We presented consistency results for graph quantizer design, where the underlying graph edit distances is generalized differentiable. As for vectors, estimators based on empirical distortion and stochastic optimization are statistically consistent. If the underlying distortion measure is a discontinuous graph edit distance, estimators based on empirical distortion measures lack theoretical justification. Thus, the proposed consistency results justify existing research on prototype-based clustering in the domain of graphs. In addition, we showed that the Lloyd-Max conditions are necessary conditions for optimality of GQ.

The mathematical framework that enables us to derive consistency results are Riemannian orbifolds. Identifying graphs with points in a Riemannian orbifold provides us locally access to a Euclidean space.  This in turn allows us to introduce geometrical and analytical concepts for extending vector quantization to the domain of graphs. The implication of this approach is that it provides us a template for consistently linking methods from structural pattern recognition other than GQ to statistical pattern recognition methods. 

\subsubsection*{Acknowledgments.}
The first author is very grateful to Vladimir Norkin for his kind support and valuable comments.

\begin{appendix}
\section{Generalized Differentiable Functions}\label{sec:GDF}
Let $\S{X} = \R^n$ be a finite-dimensional Euclidean space. A function $f: \S{X} \rightarrow \R$ is \emph{generalized differentiable} at $\vec{x}\in\S{X}$  in the sense of  Norkin \cite{Norkin1986} if there is a multi-valued map $\partial f: \S{X} \rightarrow 2^{\S{X}}$ in a neighborhood of $\vec{x}$ such that 
\begin{enumerate}
\item $\partial f(\vec{x})$ is a convex and compact set;
\item $\partial f(\vec{x})$ is upper semicontinuous at $\vec{x}$, that is, if $\vec{y}_i \to \vec{x}$ and $\vec{g}_i \in \partial f(\vec{y}_i)$ for each $i\in \N$, then each accumulation point $\vec{g}$ of $(\vec{g}_i)$ is in $\partial f(\vec{x})$;
\item for each $\vec{y} \in \S{X}$ there is a $\vec{g} \in \partial f(\vec{y})$ with $f(\vec{y}) = f(\vec{x}) + \inner{\vec{g}, \vec{y}-\vec{x}} + o\args{\vec{x}, \vec{y}, \vec{g}}$,
where 
\[
\lim_{i\to \infty} \frac{\abs{o\args{\vec{x}, \vec{y}_i, \vec{g}_i}}}{\norm{\vec{y}_i -\vec{x}}} = 0
\]
for all sequences $\vec{y}_i \to \vec{y}$ and $\vec{g}_i \to \vec{g}$ with $\vec{g}_i \in \partial f\args{\vec{y}_i}$. 
\end{enumerate}
We call $f$ \emph{generalized differentiable} if it is generalized differentiable at each point  $\vec{x}\in\S{X}$. The set $\partial f(\vec{x})$ is the \emph{subdifferential} of $f$ at $\vec{x}$ and its elements are called \emph{generalized gradients}. 

\medskip

\noindent
Generalized differentiable functions have the following properties \cite{Norkin1986}:
\begin{description}
\item[{\small(GD1)}] Generalized differentiable functions are locally Lipschitz and therefore continuous and differentiable almost everywhere.
\item[{\small(GD2)}] Continuously differentiable, convex, and concave functions are generalized differentiable.
\item[{\small(GD3)}] Suppose that $f_1, \ldots, f_n:\S{X} \rightarrow \R$ are generalized differentiable at $\vec{x}\in \S{X}$. Then 
\begin{align*}
f_*(\vec{x}) &= \min(f_1(\vec{x}), \ldots, f_m(\vec{x}))\\
f^*(\vec{x}) &= \max(f_1(\vec{x}), \ldots, f_m(\vec{x}))
\end{align*}
are generalized differentiable at $\vec{x}\in \S{X}$.
\item[{\small(GD4)}] Suppose that $f_1, \ldots, f_m:\S{X} \rightarrow \R$ are generalized differentiable at $\vec{x}\in \S{X}$ and $f_0:\R^m \rightarrow \R$ is generalized differentiable at $\vec{y} = \args{f_1(\vec{x}), \ldots, f_m(\vec{x})} \in\R^m$. Then $f(\vec{x}) = f_0(f_1(\vec{x}),\ldots, f_m(\vec{x}))$ is generalized differentiable at $\vec{x} \in \S{X}$. The subdifferential of $f$ at $\vec{x}$ is of the form
\begin{align*}
\partial f(\vec{x}) = \conv \Big\{\vec{g} \in \S{X} &:\, \vec{g} = \big[\vec{g}_1 \vec{g}_2\ldots \vec{g}_m\big]\vec{g}_0, \\
& \vec{g}_0 \in \partial f_0(\vec{y}),\\
& \vec{g}_i \in \partial f_i(\vec{x}), 1 \leq i \leq m \Big\}.
\end{align*}
where $\bracket{\vec{g}_1 \vec{g}_2\ldots \vec{g}_m}$ is a ($N\times m$)-matrix.

\item[{\small(GD5)}] Suppose that $F(\vec{x}) = \E_{\vec{z}}\bracket{f(\vec{x}, \vec{z})}$, where $f(\cdot, \vec{z})$ is generalized differentiable.  Then $F$ is generalized differentiable and its subdifferential at $\vec{x} \in \S{X}$ is of the form $\partial F(\vec{x}) = \E_{\vec{z}}\bracket{\partial f(\vec{x},\vec{z})}$.
\end{description}

\section{Proofs}

Suppose that $\S{Q} = \args{\S{X}, \Gamma, \pi}$ is a Riemannian orbifold. By $\S{U}_\delta(\vec{x}) = \cbrace{\vec{x}' \, : \, \norm{\vec{x}'} < \delta}$ we denote the open ball with center $\vec{x}$ and radius $\delta > 0$. Note that $\S{U}_\delta(\gamma(\vec{x})) = \gamma\args{\S{U}_\delta(\vec{x})}$ for all $\gamma \in \Gamma$.

\subsection{Orbifold Functions}\label{subsec:proof:orbifold-functions}

\subsection*{Continuous Orbifold Functions}
\begin{proposition}\label{prop:con}
Let $f:\S{X}_{\Gamma} \rightarrow \R$ be an orbifold function. Suppose that its lift $\tilde{f}: \S{X} \rightarrow \R$  is continuous at a vector representation $\vec{x}$ that projects to $X \in \S{X}_\Gamma$. Then $\tilde{f}$ is continuous at $\gamma(\vec{x})$ for all $\gamma \in \Gamma$. 
\end{proposition}

\proof Let $\gamma \in \Gamma$ be a permutation and $\vec{x}' = \gamma(\vec{x})$. Suppose that $(\vec{y}'_i)_{i\in \N}$ is a sequence with $\vec{y}_i' \to \vec{x}'$. Then there is a sequence $(\vec{y}_i)_{i\in\N}$ with $\gamma(\vec{y}_i) = \vec{y}_i'$ for each $i \in \N$. Since permutations are homeomorphisms, we find that 
\[
\lim_{i\to \infty}\vec{y}_i = \lim_{i\to \infty}\gamma^{-1}(\vec{y}_i') = \gamma^{-1}(\vec{x}') = \vec{x}.
\]
From continuity of $\tilde{f}$ at $\vec{x}$ follows that $\tilde{f}(\vec{y}_i) \to \tilde{f}(\vec{x})$. Since $\tilde{f}$ is invariant under group actions from $\Gamma$, we have $\tilde{f}(\vec{x}) = \tilde{f}(\vec{x}')$ and $\tilde{f}(\vec{y}_i) = \tilde{f}(\vec{y}_i')$ for each $i \in \N$. We obtain
\[
\lim_{i\to \infty}\tilde{f}\args{\vec{y}_i'} = \lim_{i\to \infty}\tilde{f}\args{\vec{y}_i} = \tilde{f}(\vec{x}) = \tilde{f}(\vec{x}').
\]
This proves that $\tilde{f}$ is continuous at each vector representation that projects to $X$.
\qed

\subsection*{Locally Lipschitz Orbifold Functions}
\begin{proposition}
Let $f:\S{X}_{\Gamma} \rightarrow \R$ be an orbifold function. Suppose that its lift $\tilde{f}: \S{X} \rightarrow \R$  is locally Lipschitz at a vector representation $\vec{x}$ that projects to $X \in \S{X}_\Gamma$. Then $\tilde{f}$ is locally Lipschitz at $\gamma(\vec{x})$ for all $\gamma \in \Gamma$. 
\end{proposition}

\proof Since $\tilde{f}$ is locally Lipschitz at $\vec{x}$ there is a $L \geq 0 $ and $\delta > 0$ such that 
\[
\abs{\tilde{f}(\vec{y})-\tilde{f}(\vec{z})} \leq L\norm{\vec{y}-\vec{z}}
\]
for all $\vec{y}, \vec{z} \in \S{U}_{\delta}(\vec{x})$. Let $\gamma \in \Gamma$ be a permutation and $\vec{x}' = \gamma(\vec{x})$. Since $\gamma$ is an isometric homeomorphism, we have
$ \S{U}_{\delta}(\vec{x}') =  \gamma\args{\S{U}_{\delta}(\vec{x})}$. From $\Gamma$-invariance of $\tilde{f}$ and the isometric property of $\gamma$ follows
\[
\abs{\tilde{f}(\vec{y}')-\tilde{f}(\vec{z}')}  = \abs{\tilde{f}(\vec{y})-\tilde{f}(\vec{z})} \leq L\norm{\vec{y}-\vec{z}} = L\norm{\vec{y}'-\vec{z}'} 
\]
for all $\vec{y}', \vec{z}' \in \S{U}_{\delta}(\vec{x}')$, where $\vec{y} = \gamma^{-1}(\vec{y}') \in  \S{U}_{\delta}(\vec{x})$ and $\vec{z} = \gamma^{-1}(\vec{z}) \in \S{U}_{\delta}(\vec{x})$.
This proves that $\tilde{f}$ is locally Lipschitz at each vector representation that projects to $X$.
\qed

\subsection*{Differentiable Orbifold Functions}

\begin{proposition}\label{prop:diff}
Let $f:\S{X}_{\Gamma} \rightarrow \R$ be an orbifold function. Suppose that its lift $\tilde{f}: \S{X} \rightarrow \R$  is differentiable at a vector representation $\vec{x}$ that projects to $X \in \S{X}_\Gamma$. Then $\tilde{f}$ is differentiable at $\gamma(\vec{x})$ for all $\gamma \in \Gamma$. The gradient of $\tilde{f}$ at $\gamma(\vec{x})$ is of the form
\[
\nabla \tilde{f}(\gamma(\vec{x})) = \gamma\args{\nabla\tilde{f}(\vec{x})}.
\]
\end{proposition}

\proof Since the lift $\tilde{f}$ of $f$ is differentiable at $\vec{x}$,  there is a $\delta > 0$ such that 
\[
\tilde{f}(\vec{x} + \vec{h}) = \tilde{f}(\vec{x}) + \inner{\nabla\tilde{f}\args{\vec{x}}, \vec{h}} + \,o(\vec{h})
\]
for all $\vec{h} \in \S{U}_\delta(\vec{0})$. Let $\vec{x}'$ be an arbitrary vector representation that projects to $X$. Then there is a $\gamma \in \Gamma$ with $\vec{x}' = \gamma(\vec{x})$. Since $\tilde{f}$ is invariant under the group actions of $\Gamma$, we have $\tilde{f}(\vec{x}') = \tilde{f}(\vec{x})$. Then for each $\vec{h}' \in \S{U}_\delta(\vec{0})$, we find that
\[
\tilde{f}(\vec{x}' + \vec{h}') - \tilde{f}(\vec{x}') =  \tilde{f}(\vec{x} + \vec{h}) - \tilde{f}(\vec{x}) =   \inner{\nabla\tilde{f}\args{\vec{x}}, \vec{h}} + \,o(\vec{h}),
\]
where $\vec{h} \in \S{X}$ with $\gamma(\vec{h}) = \vec{h}'$. Since the elements of $\Gamma$ are isometries, we have $\norm{\vec{h}} = \norm{\vec{h}'}$ giving $\vec{h} \in \S{U}_\delta(\vec{0})$. In addition, from isometry of $\gamma$ follows
\[
\inner{f_{\vec{x}}, \vec{h}} = \inner{\gamma\args{\nabla\tilde{f}\args{\vec{x}}}, \gamma(\vec{h})} = \inner{\gamma\args{\nabla\tilde{f}\args{\vec{x}}}, \vec{h}'}.
\]
We obtain
\[
\tilde{f}(\vec{x}' + \vec{h}') - \tilde{f}(\vec{x}') =  \inner{\gamma\args{\nabla\tilde{f}\args{\vec{x}}}, \vec{h}'} + \, o'(\vec{h}'),
\]
where $o'(\vec{h}') = o \circ \gamma^{-1} (\vec{h}')$ satisfies
\[
\lim_{\vec{h}' \to 0} \frac{o'(\vec{h}') }{\norm{\vec{h}'}} = \lim_{\vec{h}' \to 0} \frac{o(\gamma^{-1}(\vec{h}')) }{\norm{\vec{h}'}}  = \lim_{\vec{h}' \to 0} \frac{o(\gamma^{-1}(\vec{h}')) }{\norm{\gamma^{-1}(\vec{h}')}}  = 0.
\]
This proves that $\tilde{f}$ is differentiable at each vector representation that projects to $X$. In addition, from the proof follows that the gradient of $\tilde{f}$ at $\vec{x}' = \gamma(\vec{x})$ is of the form
\[
\nabla\tilde{f}\args{\vec{x}'} = \gamma\args{\nabla\tilde{f}\args{\vec{x}}}.
\]
\qed

\subsection*{Generalized Differentiable Orbifold Functions}

\begin{proposition}\label{prop:gendiff}
Let $f:\S{X}_{\Gamma} \rightarrow \R$ be an orbifold function. Suppose that its lift $\tilde{f}: \S{X} \rightarrow \R$  is generalized differentiable at a vector representation $\vec{x}$ that projects to $X \in \S{X}_\Gamma$. Then $\tilde{f}$ is generalized differentiable at $\gamma(\vec{x})$ for all $\gamma \in \Gamma$ and
\[
\partial \tilde{f}(\gamma(\vec{x})) = \gamma\args{\partial\tilde{f}(\vec{x})}.
\]
is a subdifferential of $\tilde{f}$ at $\gamma(\vec{x})$ for all $\gamma \in \Gamma$.
\end{proposition}

\proof Since $\tilde{f}$ is generalized differentiable at $\vec{x}$, there is a multi-valued mapping $\partial\tilde{f}: \S{U}_\delta(\vec{x})\rightarrow 2^{\S{X}}$ defined on some neighborhood $\S{U}_\delta(\vec{x})$. Let $\gamma \in \Gamma$ be an arbitrary permutation and $\vec{x}' = \gamma(\vec{x})$. Then 
\[
\partial\tilde{f}: \S{U}_\delta(\vec{x}')\rightarrow 2^{\S{X}}, \quad \vec{y}' = \gamma(\vec{y}) \mapsto \gamma\args{\partial \tilde{f}(\vec{y})}
\]
is a multi-valued mapping in a neighborhood of $\vec{x}'$. 

Since $\gamma$ is a homeomorphic linear map, we find that $\gamma(\partial\tilde{f}(\vec{x})) = \partial \tilde{f}(\vec{x}')$ is a convex and compact set. 

Next we show that $\tilde{f}$ is upper semicontinuous at $\vec{x}'$.  Suppose that $\vec{y}'_i \to \vec{x}'$, $\vec{g}'_i \in \tilde{f}_c(\vec{y}'_i)$ for each $i\in \N$, and $\vec{g}'$ is an accumulation point of $(\vec{g}'_i)_{i \in \N}$.  Then there is a $i_0 \in \N$ such that $\vec{y}'_i \in \S{U}_\delta(\vec{x}')$ for all $i \geq i_0$. From 
\[
\S{U}_\delta(\vec{x}') = \S{U}_\delta(\gamma(\vec{x})) = \gamma\args{\S{U}_\delta(\vec{x})}
\]
follows that there are vector representations $\vec{y}_i \in \S{U}_\delta(\vec{x})$ with $\gamma(\vec{y}_i) = \vec{y}_i'$ for each $i \geq i_0$. From continuity of $\gamma^{-1}$ follows that $\vec{y}_i \to \vec{x}$. By construction of $\partial \tilde{f}$ follows that
\[
\vec{g}_i' \in \partial\tilde{f}\args{\vec{y}_i'} = \partial\tilde{f}\args{\gamma\args{\vec{y}_i}} = \gamma\args{\partial\tilde{f}\args{\vec{y}_i}}
\]
for each $i\geq i_0$. Hence, there are vector representations $\vec{g}_i \in \partial\tilde{f}(\vec{y}_i)$ with $\gamma(\vec{g}_i) = \vec{g}_i'$ for each $i \geq i_0$. Since $\tilde{f}$ is upper semicontinuous at $\vec{x}$, we find that $\vec{g} \in \partial\tilde{f}(\vec{x})$. Again by construction of $\partial \tilde{f}$ follows that  
\[
\vec{g}' = \gamma(\vec{g}) \in \gamma\args{\partial\tilde{f}(\vec{x})} = \partial\tilde{f}\args{\gamma(\vec{x})} = \partial \tilde{f}(\vec{x}').
\]
This proves upper semicontinuity of $\partial \tilde{f}$ at all vector representations projecting to $X = \pi(\vec{x})$. 

Finally, we prove that $\tilde{f}$ satisfies the subderivative property at $\vec{x}'$. Suppose that $\vec{y}', \vec{y} \in \S{X}$ with $\vec{y}' = \gamma(\vec{y})$. By $\Gamma$-invariance of $\tilde{f}$, we have $\tilde{f}(\vec{y}') = \tilde{f}(\vec{y})$. Since $\tilde{f}$ is generalized differentiable at $\vec{x}$, we find a $\vec{g} \in \partial \tilde{f}(\vec{y})$ such that
\[
\tilde{f}(\vec{y}') = \tilde{f}(\vec{y}) = \tilde{f}(\vec{x}) + \inner{\vec{g}, \vec{y}-\vec{x}} +\, o(\vec{x}, \vec{y}, \vec{g})
\]
with $o(\vec{x}, \vec{y}, \vec{g})$ tending faster to zero than $\norm{\vec{y}-\vec{x}}$. Let $\vec{g}' = \gamma(\vec{g})$. Exploiting $\Gamma$-invariance of $\tilde{f}$ as well as isometry and linearity of $\gamma$ yields
\begin{align*}
\tilde{f}(\vec{y}') &= \tilde{f}(\gamma(\vec{x})) + \inner{\gamma(\vec{g}),\gamma(\vec{y}-\vec{x})} +\, o(\vec{x}, \vec{y}, \vec{g})\\
&= \tilde{f}(\vec{x}') + \inner{\vec{g}',\vec{y}'-\vec{x}'} +\, o(\vec{x}, \vec{y}, \vec{g}).
\end{align*}
We define $o'(\vec{x}', \vec{y}', \vec{g}') = o\circ \gamma^{-1}(\vec{x}', \vec{y}', \vec{g}') = o(\vec{x}, \vec{y}, \vec{g})$ showing that $o'$ tends faster to zero than $norm{\vec{y}' -\vec{x}}$. This proves the subderivative property of $\tilde{f}$  at all vector representations projecting to $X = \pi(\vec{x})$. 

Putting all results together yields that $\tilde{f}$ is generalized differentiable at $\gamma(\vec{x})$ for all $\gamma \in \Gamma$.
\qed

\subsection{Lloyd-Max Necessary Conditions for Optimality}\label{subsec:NCGQ}

Due to the comparable nice analytical properties of Riemannian orbifolds, the proofs for the nearest neighbor and centroid condition of optimal graph quantizers are similar to their respective counterparts in vector quantization. 

\begin{theorem}[Nearest Neighbor Condition]\label{theorem:NNC}
Suppose that $\S{C}$ is a fixed codebook. Any graph quantizer $Q:\S{X}_\Gamma \rightarrow \S{C}$ with 
\[
Q(X) = \arg\min_{Y \in \S{C}}  d\args{X, Y}
\]
for all $X\in \S{X}_\Gamma$, where ties are resolved according to some rule, has minimal expected distortion.
\end{theorem}

\proof Suppose that $Q':\S{X}_\Gamma \rightarrow \S{C}$ is a graph quantizer with arbitrary regions. Then we have
\[
d(X,Q'(X)) \geq \min_{Y \in \S{Y}} d(X, Y) = d(X, Q(X))
\]
for all $X \in \S{X}_\Gamma$. This implies
\[
D(Q')= \E_{X} \bracket{d\args{X, Q'(X)}} \geq \E_{X} \bracket{d\args{X, Q(X)}} = D(Q).
\]
\qed

\begin{theorem}[Nearest Neighbor Condition]\label{theorem:CC}
Suppose that $\S{P}_Q$ is a fixed partition and $Q:\S{X}_\Gamma \rightarrow \S{C}$ a graph quantizer with codebook $\S{C}$ satisfying
\[
Y_j = \arg\min_{Y\in \S{X}_\Gamma} \E\bracket{d\args{X, Y}\,|\, X\in\S{R}_j}
\]
for all $Y\in \S{X}_\Gamma$ and all $j \in \S{J}$. Then $Q$ has minimal expected distortion.
\end{theorem}

\proof Let $P_j = P(X \in \S{R}_j)$. Suppose that $Q'$ is a quantizer with partition $\cbrace{\S{R}_1, \ldots, \S{R}_k}$ and arbitrary codebook $\S{C} = \cbrace{Y_1', \ldots, Y_k'}$.
Then we have
\begin{align*}
\E\bracket{d(X,Q'(X))} &= \sum_{j=1}^k P_j \E\bracket{d(X, Q'(X))\,|\, X \in \S{R}_j}\\
&= \sum_{j=1}^k P_j \E\bracket{d(X, Y_j')\,|\, X \in \S{R}_j}\\
&\geq \sum_{j=1}^k P_j  \min_{Y \in \S{X}_\Gamma}\E\bracket{d(X, Y)\,|\, X \in \S{R}_j}\\
&= \sum_{j=1}^k P_j  \E\bracket{d(X, Y_j)\,|\, X \in \S{R}_j} = \E\bracket{d(X,Q(X))}
\end{align*}
\qed

\end{appendix}
\bibliographystyle{splncs}

\end{document}